\def\arxivversion{1}
\PassOptionsToPackage{table}{xcolor}
\documentclass{article}
\usepackage{iclr2027_conference,times}

\usepackage{graphicx}
\usepackage{hyperref}
\hypersetup{hidelinks}
\usepackage{url}
\usepackage{caption}
\usepackage{placeins}
\usepackage{flafter}

\usepackage{algorithm}
\usepackage{algorithmic}

\usepackage{booktabs}

\usepackage{amsmath}
\usepackage{amssymb}
\usepackage{amsfonts}
\usepackage{tikz}

\definecolor{softblue}{RGB}{219,234,254}
\definecolor{softgreen}{RGB}{22,141,71}
\definecolor{sectiongray}{RGB}{225,225,225}
\definecolor{ourspink}{RGB}{255,228,236}

\newcommand{\equalmark}{\textsuperscript{$\diamondsuit$}}
\newcommand{\corrmark}{\textsuperscript{$\dagger$}}
\newcommand{\fleaf}{%
  \raisebox{0.75ex}{%
    \tikz[scale=0.065,baseline=-0.4ex]{%
      \fill[black]
        (0,0) .. controls (1.25,0.05) and (1.85,1.15) .. (1.75,2.25)
        .. controls (0.75,2.05) and (-0.20,1.15) .. cycle;
      \draw[black,line width=0.45pt] (0,-0.15)--(1.25,1.65);
    }%
  }%
}

\title{DeepVoyager-VL: Incentivizing Vision-in-the-Loop Search for Long-Horizon Multimodal Agents}

\ifdefined\arxivversion
  \author{%
    \begin{minipage}{0.95\textwidth}
    \centering
    \small\bfseries
    Huanyao Zhang\textsuperscript{1}\equalmark\fleaf\hspace{0.7em}
    Jiepeng Zhou\textsuperscript{2}\equalmark\hspace{0.7em}
    Runhao Zhao\textsuperscript{3}\equalmark\hspace{0.7em}
    Yanzhe Shan\textsuperscript{4}\hspace{0.7em}
    Jiaoyang Chen\textsuperscript{5}\\[2pt]
    Bowen Zhou\textsuperscript{1}\quad
    Bo Li\textsuperscript{1}\quad
    Fang Wang\textsuperscript{1}\quad
    Jialong Wu\textsuperscript{1}\quad
    Zhengwei Tao\textsuperscript{1}\\[2pt]
    Lang Mei\textsuperscript{6}\quad
    Xiaohan Yu\textsuperscript{6}\quad
    Liyan Liu\textsuperscript{6}\quad
    Chong Chen\textsuperscript{6}\corrmark\quad
    Wentao Zhang\textsuperscript{1}\corrmark\\[5pt]
    \normalsize\normalfont
    \textsuperscript{1}PKU\quad
    \textsuperscript{2}HKUST(GZ)\quad
    \textsuperscript{3}NUDT\quad
    \textsuperscript{4}OUC\quad
    \textsuperscript{5}HITSZ\quad
    \textsuperscript{6}Huawei Cloud BU\\[4pt]
    \small
    $\diamondsuit$ Equal contribution\quad
    \fleaf Project Leader\quad
    $\dagger$ Corresponding authors
    \end{minipage}
  }
  \iclrfinalcopy
\else
  \author{Anonymous Authors}
\fi

\begin{document}

\maketitle

\ifdefined\arxivversion
  \fancyhead{}
  \lhead{Preprint}
\fi

\begin{abstract}
Multimodal large language models (MLLMs) have advanced visual understanding and reasoning, yet their static parametric knowledge limits their ability to address knowledge-intensive and dynamically evolving open-world problems. To move beyond this limitation, multimodal deep search has emerged as a key direction for open-world information access, evolving from single-turn factual retrieval toward long-horizon, multi-turn search guided by visual evidence. However, existing methods typically confine vision to the input or answer stage, overlooking its role in intermediate reasoning, and lack designs tailored to long-horizon interaction. Consequently, visual evidence rarely drives continued retrieval, constraining both interaction depth and reasoning span. To address these limitations, we propose \textbf{DeepVoyager-VL}, a long-horizon multimodal deep-search framework for vision-in-the-loop search. Specifically, we construct a multimodal event graph to drive data synthesis, yielding problems with intermediate visual dependencies and long reasoning chains. We then design an agent framework for active visual acquisition and on-demand image loading. Finally, we fine-tune models on the synthesized data without reinforcement learning. Extensive experiments across ten multimodal search benchmarks demonstrate the effectiveness of our method.
\end{abstract}


\section{Introduction}

\begin{figure}[t]
    \centering
    \includegraphics[width=\linewidth]{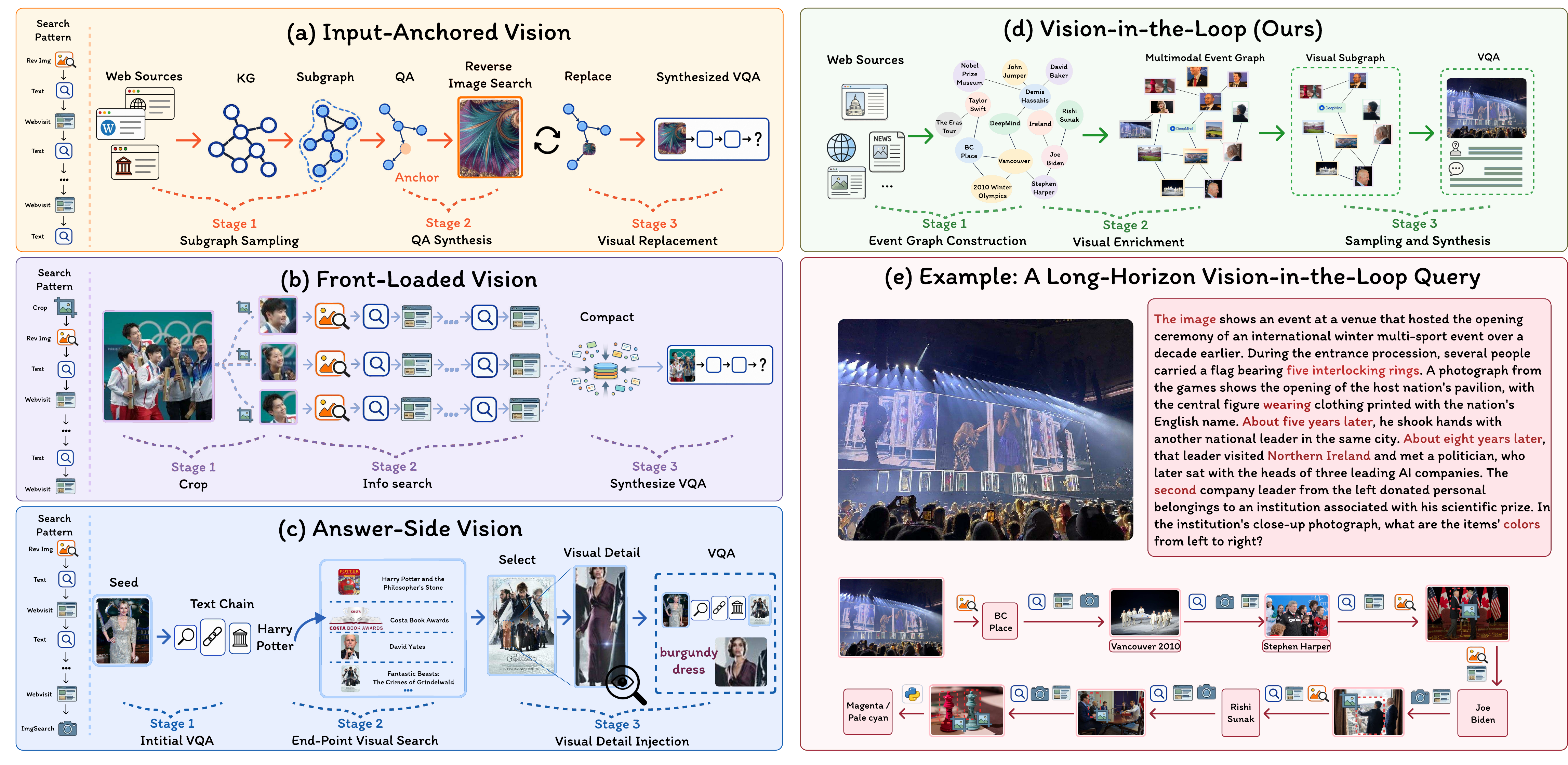}
    \caption{Comparison of multimodal search data synthesis paradigms. Prior methods place vision at the input through entity substitution (a), concentrate visual reasoning before text-based search (b), or graft visual evidence near the answer (c). We instead synthesize vision-in-the-loop questions from a visually enriched multimodal event graph (d), as illustrated by a representative long-horizon example (e).}
    \label{fig:teaser}
\end{figure}

Multimodal large language models (MLLMs) have made remarkable progress in visual understanding and reasoning~\citep{llava,qwen25vl,internvl3}, yet their fixed parametric knowledge limits them on knowledge-intensive and dynamically evolving problems. Deep-search agents address this limitation by gathering external evidence through iterative reasoning and tool use~\citep{react,searchr1,webdancer,websailor}, and recent systems extend the paradigm with image search and visual tools~\citep{mmsearchr1,webwatcher,visiondeepresearch,opensearchvl}. Real-world investigations, however, often require linking evidence scattered across pages, modalities, events, and time. An image discovered mid-search may resolve an intermediate entity that determines the next query, while the final answer emerges only after several such transitions. Agents must therefore use newly acquired visual evidence to drive retrieval and sustain this process over long horizons---a paradigm we call \textbf{vision-in-the-loop search}. Benchmarks increasingly reflect this shift: early ones reduce vision to identifying the input image before textual retrieval~\citep{mmsearch,livevqa}; later ones require finer input-side perception~\citep{mmsearchplus,vdrbench}; and recent evaluations require inspecting newly discovered images and continuing search from visual observations~\citep{mmbrowsecomp,browsecompv3,visbrowse}. The frontier is thus moving from one-shot perception to extended search guided by acquired visual evidence.

Existing methods are not yet equipped for this shift, for two related reasons. First, current training data provide little supervision for intermediate visual dependencies. As illustrated in Figure~\ref{fig:teaser}(a)--(c), representative paradigms place vision at the input through entity substitution~\citep{webwatcher,opensearchvl}, concentrate it before largely textual search~\citep{visiondeepresearch}, or graft it near the answer~\citep{visualseeker}. In each case, the reasoning chain is constructed in text first, so an acquired visual observation rarely becomes necessary for a later retrieval action. Second, sustaining such transitions over long horizons creates an evidence-management challenge: loading every candidate image incurs costly multimodal context, while retaining only textual summaries may discard details needed later. Existing mechanisms manage context retention or compression~\citep{lmmsearcher,pointsseeker,simplesearchvl}, but do not fully connect candidate discovery, selective observation, local visual operations, and follow-up retrieval across turns. Thus, existing data do not teach the loop, and existing agents struggle to sustain it.

To bridge this gap, we propose \textbf{DeepVoyager-VL}, a long-horizon multimodal deep-search framework that constructs and learns vision-in-the-loop behavior. We first introduce \textbf{EventVoyage-VL}, which organizes real-world events through temporal, spatial, and participant associations, enriches them with retrievable visual evidence, and composes explicit inference structures before realizing them as questions. This structure-before-language process makes intermediate visual evidence necessary for subsequent retrieval, as shown in Figure~\ref{fig:teaser}(d)--(e). We then stratify problem difficulty and extract multi-turn teacher trajectories under a unified interaction protocol. Image discovery yields lightweight, referenceable candidates, while active acquisition selectively materializes relevant images or crops and preserves them for later actions. Finally, we distill the curated trajectories through supervised fine-tuning alone, without an additional reinforcement-learning stage.

Our main contributions are summarized as follows:
\begin{itemize}
    \item We introduce \textbf{EventVoyage-VL}, a structure-before-language synthesis pipeline built on a visually enriched multimodal event graph, to construct long-horizon problems with explicit intermediate visual dependencies.
    \item We develop \textbf{DeepVoyager-VL}, a long-horizon search agent that actively acquires referenceable visual evidence and learns vision-in-the-loop behavior from supervised trajectories alone.
    \item Extensive experiments across ten multimodal search benchmarks demonstrate the effectiveness of DeepVoyager-VL across diverse task settings.
\end{itemize}

\section{Related Work}

\subsection{Multimodal Large Language Models}

Multimodal large language models (MLLMs) have advanced visual understanding through large-scale cross-modal pretraining~\citep{llava,llavaonevision,qwen25vl,internvl3,gpt4o,gemini25,qwen3vl,glm45v} and post-training techniques that strengthen reasoning and enable active image manipulation~\citep{deepseekr1,grpo,thinkingwithimages,twisurvey,deepeyes,minio3,pixelreasoner,pyvision,thyme}. However, their evidence remains bounded by parametric knowledge and images supplied at inference. Knowledge-intensive, evolving open-world problems exceed both boundaries, creating a clear need to acquire current evidence through external search.

\subsection{Multimodal Deep Search Agents}

Deep-search agents iteratively reason and use tools to acquire information~\citep{react,searchr1,webdancer,websailor,tongyidr}. Multimodal variants add visual tools and learn search policies through supervised fine-tuning and outcome-based reinforcement learning~\citep{mmsearchr1,webwatcher,visiondeepresearch,opensearchvl,skyworkr1v4,visualseeker}. Yet real-world needs are shifting from single-turn factual retrieval to long-horizon, cross-page and cross-modal search, where visual evidence repeatedly guides subsequent queries. Recent systems control context growth by offloading visual assets~\citep{lmmsearcher} or folding stale context into visual space~\citep{pointsseeker}. These effective mechanisms passively manage evidence after it enters context; we argue that actively deciding when and what to load is equally important. We therefore decouple image discovery from visual observation, enabling agents to select relevant images, load them on demand, and use them for subsequent retrieval.

\begin{figure}[t]
  \centering
  \includegraphics[width=0.95\linewidth]{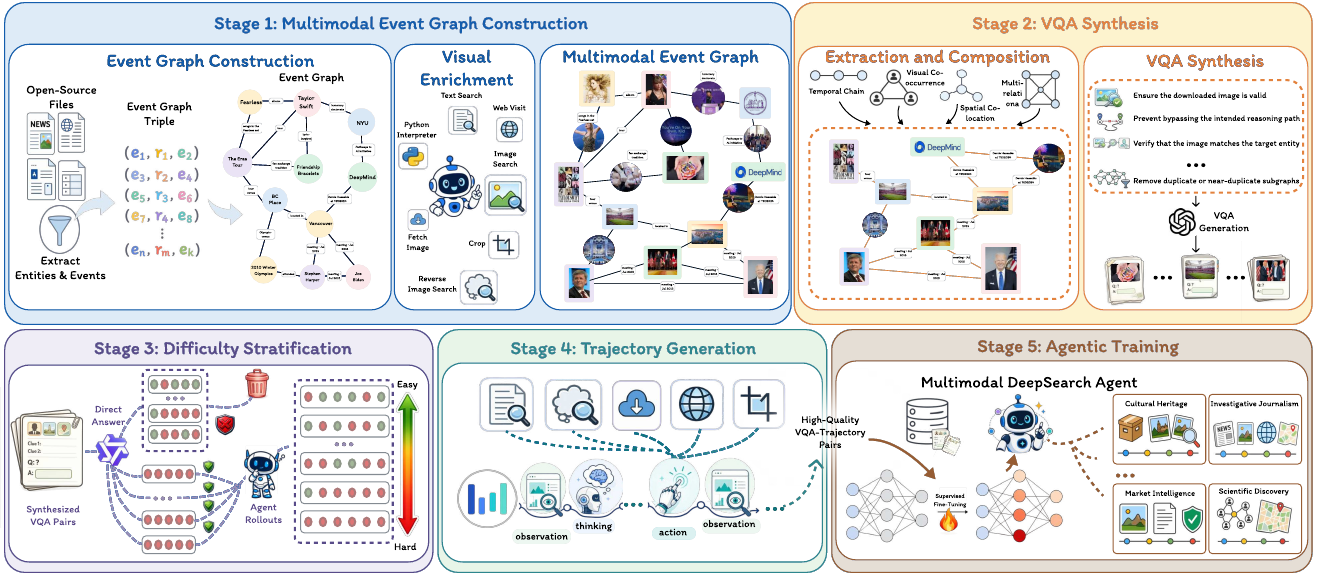}
    \caption{Overview of DeepVoyager-VL, encompassing vision-in-the-loop data synthesis, difficulty-aware trajectory curation, and supervised agent training.}
    \label{fig:pipeline}
\end{figure}

\subsection{Datasets and Benchmarks}

Multimodal search benchmarks increasingly move vision deeper into the reasoning process. Early datasets use an input image mainly to identify an entity~\citep{mmsearch,livevqa}, while later ones require finer input-side perception~\citep{mmsearchplus,vdrbench}. Recent benchmarks further probe visual reasoning during search, emphasizing newly discovered visual information, cross-modal integration, fine-grained perception, and interleaved multimodal agentic search~\citep{visbrowse,browsecompv3,hou2026interlv}. Training-data synthesis has not fully followed this shift. Entity substitution replaces a mention in a text-built chain with an image~\citep{webwatcher,opensearchvl}; front-loaded cropping performs visual exploration before a text-only trajectory~\citep{visiondeepresearch}; and answer-side grafting adds a visual retrieval-and-perception subchain near the target~\citep{visualseeker}. Despite their different insertion points, all three construct the reasoning chain in text first, so intermediate visual observations do not determine subsequent hops. We instead synthesize from a multimodal event graph whose spatio-temporal and causal structure supports visual dependencies throughout long-horizon questions.

\section{Method}

\subsection{Overview and Problem Formulation}

Multimodal deep research differs from text-only research when newly acquired visual evidence determines what to search next. An image may reveal an entity, inscription, or scene absent from surrounding text; once this observation resolves an intermediate variable and instantiates a later query, vision becomes a driver rather than an input-only or terminal modality. We call this property \emph{vision in the loop}. Figure~\ref{fig:pipeline} outlines the five-stage pipeline: data synthesis (Stages~1--2), difficulty-aware trajectory curation (Stages~3--4), and supervised agent training (Stage~5; Appendix~\ref{app:method-details}).

Given a query $x=(q,\mathcal{I}_s)$, an agent interacts with an open-web environment $\mathcal{E}$. At turn $t$, it generates reasoning $r_t$ and action $a_t$, receives observation $o_t$, and updates its context:
\begin{equation}
\begin{aligned}
(r_t,a_t)&\sim\pi_\theta(\cdot\mid C_t),\qquad o_t=\mathcal{E}(a_t),\\
C_{t+1}&=C_t\oplus(r_t,a_t,o_t).
\end{aligned}
\label{eq:interaction}
\end{equation}
The resulting trajectory is $\tau=(x,r_1,a_1,o_1,\ldots,r_T,a_T,o_T,y)$. Let $o_t^{\mathrm v}\leadsto a_{t'}$ mean that a newly acquired visual observation resolves a variable required by a later action. Then $\tau$ is vision-in-the-loop if
\begin{equation}
\exists\,1\le t<t'\le T:\quad o_t^{\mathrm v}\leadsto a_{t'}.
\label{eq:vil}
\end{equation}
To exclude input-only or terminal vision, we also require counterfactual visual necessity. Let $\mathcal{Y}(q,\mathcal{I}_s,\mathcal{I}_r,\mathcal{K})$ denote the answers consistent with retrieved images $\mathcal{I}_r$ and other evidence $\mathcal{K}$:
\begin{equation}
\mathcal{Y}(q,\mathcal{I}_s,\mathcal{I}_r,\mathcal{K})=\{y\},\qquad
|\mathcal{Y}(q,\mathcal{I}_s,\varnothing,\mathcal{K})|>1.
\label{eq:visual-necessity}
\end{equation}
Together, Equations~\ref{eq:vil} and~\ref{eq:visual-necessity} require vision to occur within the inference program and be necessary for the answer.

\subsection{Vision-in-the-Loop Data Synthesis}

Direct generation often produces decorative images, pseudo-hops, or image--text mismatches. We instead separate deterministic program construction from constrained QA realization: the former fixes the evidence dependency and answer, while the latter verbalizes the verified program.

\paragraph{Multimodal event graph construction and enrichment.}
We extract entities and events from Wikipedia and multi-domain news, normalizing each record as $r=(u^s,u^t,c,t,l)$ with participants, relation type, time, and location. Spatio-temporally co-occurring records form macro-events, and repeated entity-pair interactions form relation bundles, yielding $\mathcal{G}_0=(\mathcal{V},\mathcal{E},\mathcal{M})$.

\noindent\textbf{\textit{Agentic visual evidence enrichment.}}
A search agent adds images serving as input anchors, cross-entity transitions, or visual intermediates and endpoints, with soft balancing across domains. Each image $I$ stores \emph{retrieval context} $K_{\mathrm{ret}}(I)$ for relocating it and \emph{visual-content evidence} $F_{\mathrm{vis}}(I)$ for image-grounded attributes, counts, text, objects, and relations. We prevent answer leakage by enforcing
\begin{equation}
K_{\mathrm{ret}}(I)\cap\operatorname{Alias}(y)=\varnothing.
\label{eq:anti-leakage}
\end{equation}
Identity hypotheses from visual recognition or contextual search are cross-checked against page and image evidence. The enriched graph is
\begin{equation}
\begin{aligned}
\mathcal{G}&=(\mathcal{V},\mathcal{E},\mathcal{M},\mathcal{I},\mathcal{A},\mathcal{F}),\\
\mathcal{F}&=\{(I,K_{\mathrm{ret}}(I),F_{\mathrm{vis}}(I)):I\in\mathcal{I}\},
\end{aligned}
\label{eq:mm-graph}
\end{equation}
where $\mathcal{A}$ aligns images with entities or events; all graph and image fields retain provenance.

\paragraph{Subgraph extraction, composition, and VQA synthesis.}
For a structural and modality predicate $\rho$, we deterministically extract connected subgraphs
\begin{equation}
\Phi_\rho(\mathcal{G})=\{g\subseteq\mathcal{G}:\rho(g)=1\}.
\end{equation}
The extracted structures comprise \emph{Visual Co-occurrence Networks} over shared images or events, \emph{Temporal Event Chains} over common participants, \emph{Spatial Co-location Structures} linking scene-grounded places to co-located events, and \emph{Multi-relational Dense Motifs} combining convergent constraints. Visual nodes orthogonally provide \emph{Scene}, \emph{Object-centric}, or \emph{Visual Knowledge Grounding}; their composition turns local information into a global dependency.

Let $\operatorname{in}(g_i)$ and $\operatorname{out}(g_i)$ denote the input and output interface variables of primitive $g_i$. A program $g_1\circ\cdots\circ g_n$ requires compatible adjacent interfaces and at least one visually resolved internal interface:
\begin{equation}
\begin{aligned}
\operatorname{out}(g_i)&=\operatorname{in}(g_{i+1}),&&i<n,\\
\exists j<n:\quad \operatorname{out}(g_j)&\in\operatorname{Var}(F_{\mathrm{vis}}(\mathcal{I}_r)).
\end{aligned}
\label{eq:composition}
\end{equation}
The downstream target must also be absent from the source images to prevent bypass. Each subgraph becomes an inference program $\Pi=(\mathcal{D},U,\mathcal{L},\mathcal{I}_s,\mathcal{I}_r,y)$ with reasoning DAG $\mathcal{D}$, candidate universe $U$, and constraints $\mathcal{L}$, and is built backward from a visual endpoint under
\begin{equation}
|U_{\mathcal{L}}|=1,\qquad
|U_{\mathcal{L}\setminus\{\ell\}}|>1\quad\forall\ell\in\mathcal{L},
\label{eq:minimal-constraints}
\end{equation}
so every principal constraint is necessary within candidate universe $U$.

\paragraph{QA realization and quality control.}
A multimodal generator verbalizes $\Pi$ with canonical names, then applies controlled masking to obtain the final question paired with $y$. Visual claims use only $F_{\mathrm{vis}}$, while provenance-backed text forms retrieval bridges. Validators check program completeness, constraint minimality, leakage, duplicates, image--text alignment, and evidence readability. Counterfactual removal tests solvability and visual necessity, and bilingual realizations are checked for semantic equivalence.

\subsection{Long-Horizon Multimodal Search Agent}

Long interactions dilute relevant evidence, while eagerly loaded images consume disproportionate context. We therefore make context growth follow evidence need rather than tool-return volume.

\paragraph{Agent framework and graded evidence interface.}
DeepVoyager-VL follows Equation~\ref{eq:interaction}, executing one tool action per turn with optional batching within a call. Its actions support candidate discovery (\textsc{TextSearch}, \textsc{ImageSearch}, \textsc{ReverseImageSearch}), goal-directed reading (\textsc{WebVisit}), active perception (\textsc{FetchImage}, \textsc{CropImage}), and computation (\textsc{PythonInterpreter}). Search metadata provide navigation clues; visual claims require an explicitly loaded image.

Before outputs from \textsc{WebVisit}, \textsc{ImageSearch}, or \textsc{ReverseImageSearch} enter policy context, a helper vision-language model retains only goal-relevant conclusions, source cues, and image references. These query-dependent summaries guide navigation but do not replace answer-bearing images.

\paragraph{Active visual acquisition.}
We separate image \emph{discoverability} from \emph{observability}. Search exposes lightweight URL--caption references; \textsc{FetchImage} materializes a selected image, and \textsc{CropImage} creates a reusable local observation. Thus, candidates add only references, while visual tokens enter policy context through explicit fetch or crop actions.

Algorithm~\ref{alg:agent} gives the complete loop. Visual register $V$ preserves observations across turns, and tool errors remain recoverable observations. Let $\mathcal{T}_{\mathrm v}$ contain the two visual tools and $\mathcal{T}_{\mathrm g}$ the three goal-conditioned tools above.

\begin{algorithm}[t]
\caption{DeepVoyager-VL agent interaction loop}
\label{alg:agent}
\begin{algorithmic}[1]
\REQUIRE Query $x=(q,\mathcal{I}_s)$, policy $\pi_\theta$, budget $T$
\STATE $C\leftarrow x$; $V\leftarrow\mathcal{I}_s$
\FOR{$t=1$ to $T$}
    \STATE $(r,a)\sim\pi_\theta(\cdot\mid C)$
    \IF{$a=\textsc{Answer}(y)$}
        \RETURN $y$
    \ENDIF
    \STATE $o\leftarrow\mathcal{E}(a)$
    \IF{$a\in\mathcal{T}_{\mathrm v}$}
        \STATE $V\leftarrow V\cup\{o\}$
    \ELSIF{$a\in\mathcal{T}_{\mathrm g}$}
        \STATE $o\leftarrow\operatorname{GoalExtract}(o)$
    \ENDIF
    \STATE $C\leftarrow C\oplus(r,a,o)$
\ENDFOR
\RETURN $\operatorname{ForceAnswer}(\pi_\theta,C)$
\end{algorithmic}
\end{algorithm}

\subsection{Data Curation and Agent Training}

\paragraph{Difficulty stratification and teacher trajectory generation.}
A direct-answer probe retains questions consistently failed without tools, while tool-augmented rollouts stratify them as easy, medium, or hard by empirical Pass@$K$. These rollouts measure difficulty only. Under the agent interface above, a stronger teacher then regenerates a trajectory for every retained question; an LLM judge and protocol validators retain correct, legal, and replayable trajectories. Separating difficulty estimation from trajectory generation prevents noisy probe behavior from entering the supervision.

\paragraph{Supervised fine-tuning.}
The curated corpus combines general search and vision-in-the-loop trajectories, serialized under a unified multi-turn protocol as $z_i=(z_{i,1},\ldots,z_{i,L_i})$. This shared format aligns tool semantics across data sources. We optimize
\begin{equation}
\mathcal{L}_{\mathrm{SFT}}(\theta)=
-\frac{\sum_i\sum_{t=1}^{L_i}m_{i,t}\log p_\theta(z_{i,t}\mid z_{i,<t})}
{\sum_i\sum_{t=1}^{L_i}m_{i,t}},
\label{eq:sft}
\end{equation}
where $m_{i,t}=1$ only for assistant reasoning, tool calls, and answers; all other messages and observations are conditioning context. We update the language backbone while freezing the visual encoder and multimodal merger, without reinforcement learning.


\setlength{\tabcolsep}{6pt} %

\ifdefined\mainresultinline
\begin{minipage}{\textwidth}
\captionsetup{type=table}
\else
\begin{table}[t]
\fi
\caption{Performance across ten multimodal information-seeking benchmarks. $\Delta$ is the absolute-point gain over the corresponding Agentic Workflow base model. Within each group, the best and second-best results are bolded and underlined, respectively. \textsuperscript{$\dagger$}~denotes initialization from a Thinking checkpoint; \textsuperscript{$\ddagger$}~denotes LMM-Searcher's 30-turn\,/\,100-turn settings ($x$\,/\,$y$).}
\label{tab:my-table-2-test}
\centering
\footnotesize
\resizebox{\linewidth}{!}{
\begin{tabular}{l|cccccccccc|c}
\toprule[1.2pt]
\textbf{Model} & \textbf{MMSearch} & \textbf{SimpleVQA} & \textbf{LiveVQA} & \textbf{FVQA} & \textbf{BC-VL} & \textbf{MM-BC} & \textbf{MMSearch+} & \textbf{VDR} & \textbf{BC-V\textsuperscript{3}} & \textbf{VisBrowse} & \textbf{Avg.} \\
\midrule
\rowcolor{sectiongray}\multicolumn{12}{c}{\rule{0pt}{2.6ex}\textbf{Direct Answer}\rule[-1.0ex]{0pt}{0pt}} \\

\midrule
GPT-5.5 & \textbf{68.7} & \underline{67.0} & \textbf{73.0} & \textbf{66.7} & \textbf{47.9} & \textbf{17.5} &
\underline{20.3} & \textbf{18.6} & \textbf{23.0} & \textbf{36.1} & \textbf{43.9} \\
Gemini-3.1-Pro & \underline{64.2} & 64.1 & \underline{65.0} & 58.9 & 41.4 & 11.5 & \textbf{26.4} & \underline{15.6} &
\underline{19.3} & 23.7 & \underline{39.0} \\
Claude-Opus-4.6 & 59.8 & \textbf{71.7} & 53.1 & \underline{60.1} & \underline{43.5} & \underline{13.2} & 13.2 & 15.4 & 15.0 & \underline{27.2}
& 37.2 \\
Qwen3-VL-30B-A3B-Instruct & 18.7 & 53.2 & 42.7 & 34.7 & 29.6 & 4.0 & 3.2 & 3.8 & 6.0 & 11.2 &
20.7 \\
Qwen3-VL-8B-Instruct & 15.2 & 44.7 & 41.0 & 28.0 & 25.1 & 4.9 & 3.2 & 2.8 & 1.0 & 8.9 &
17.5 \\
\midrule

\rowcolor{sectiongray}\multicolumn{12}{c}{\rule{0pt}{2.6ex}\textbf{Agentic Workflow}\rule[-1.0ex]{0pt}{0pt}} \\
\midrule
GPT-5.5 & \textbf{82.7} & \textbf{82.3} & \textbf{90.3} & \textbf{84.3} & \textbf{68.2} & \textbf{51.9} & 48.6 & \textbf{42.0} & \textbf{55.0} & \textbf{66.3} & \textbf{67.2} \\
Gemini-3.1-Pro & \textbf{82.7} & 81.0 & 87.3 & \underline{81.3} & \underline{65.9} & 44.4 & \underline{51.5} & \underline{40.8} & \underline{50.0} & 62.1 & 64.7 \\
Claude-Opus-4.6 & \underline{81.7} & \underline{81.7} & \underline{88.0} & 79.3 & 63.2 & \underline{49.4} & \textbf{52.1} & 36.6 & \textbf{55.0} & \underline{62.7} & \underline{65.0} \\
Qwen3-VL-30B-A3B-Instruct & 64.7 & 71.0 & 73.3 & 72.3 & 41.6 & 9.9 & 17.7 & 21.6 & 11.3 & 23.1 & 40.7 \\
Qwen3-VL-8B-Instruct & 61.7 & 60.0 & 67.7 & 69.7 & 34.6 & 5.8 & 13.5 & 16.8 & 7.7 & 15.4 & 35.3 \\

\midrule
\rowcolor{sectiongray}\multicolumn{12}{c}{\rule{0pt}{2.6ex}\textbf{Multimodal Deep Search Agents}\rule[-1.0ex]{0pt}{0pt}} \\
\midrule
\rowcolor{sectiongray!55}\multicolumn{12}{l}{\textbf{\textit{Small-scale}}} \\
\midrule
MMSearch-R1-7B & 53.8 & 57.4 & 48.4 & 58.4 & -- & -- & -- & -- & 4.0 & -- & -- \\
WebWatcher-7B & 49.1 & 54.3 & 51.2 & -- & 21.2 & -- & -- & -- & 4.7 & -- & -- \\
DeepEyesV2-7B & 63.7 & 59.4 & -- & 60.6 & -- & -- & -- & -- & -- & -- & -- \\
SenseNova-MARS-8B & 67.8 & 70.2 & 56.2 & 67.1 & -- & -- & -- & -- & -- & -- & -- \\
Vision-DeepResearch-8B & 69.6 & -- & 76.7 & 64.7 & 42.6 & -- & 20.4 & \underline{29.2} & \underline{11.7} & -- & -- \\
MM-DeepResearch-8B & 67.8 & 65.9 & 65.0 & 69.2 & 37.9 & -- & -- & -- & -- & -- & -- \\
POINTS-Seeker-8B & 70.8 & 68.8 & \underline{77.7} & 71.2 & 44.4 & -- & 25.2 & -- & -- & -- & -- \\
OpenSearch-VL-8B & 64.5 & 71.6 & 59.6 & 71.5 & 37.6 & -- & -- & 20.8 & -- & -- & -- \\
SimpleSearch-VL-8B & \textbf{77.1} & \textbf{76.6} & 75.2 & \underline{76.8} & \underline{52.1} & -- & \underline{32.5} & -- & -- & -- & -- \\
Visual-Seeker-8B & 72.2 & -- & -- & -- & 47.6 & \underline{16.1} & 27.3 & -- & -- & \underline{34.7} & -- \\
\rowcolor{ourspink}\textbf{DeepVoyager-VL-8B} & \underline{72.7} & \underline{76.3} & \textbf{82.7} & \textbf{82.7} & \textbf{58.4} & \textbf{24.0} & \textbf{37.1} & \textbf{35.0} & \textbf{32.3} & \textbf{47.3} & \textbf{54.8} \\
\rowcolor{ourspink!45}\textcolor{softgreen}{\textbf{$\Delta$ vs. base}} & \textcolor{softgreen}{\textbf{+11.0}} & \textcolor{softgreen}{\textbf{+16.3}} & \textcolor{softgreen}{\textbf{+15.0}} & \textcolor{softgreen}{\textbf{+13.0}} & \textcolor{softgreen}{\textbf{+23.8}} & \textcolor{softgreen}{\textbf{+18.2}} & \textcolor{softgreen}{\textbf{+23.6}} & \textcolor{softgreen}{\textbf{+18.2}} & \textcolor{softgreen}{\textbf{+24.6}} & \textcolor{softgreen}{\textbf{+31.9}} & \textcolor{softgreen}{\textbf{+19.5}} \\
\midrule
\rowcolor{sectiongray!55}\multicolumn{12}{l}{\textbf{\textit{Large-scale}}} \\
\midrule
WebWatcher-32B & 55.3 & 59.0 & 58.7 & -- & 27.0 & -- & -- & -- & \underline{8.7} & -- & -- \\
SenseNova-MARS-32B & \underline{74.3} & 74.1 & 60.8 & 72.6 & -- & -- & -- & -- & -- & -- & -- \\
Skywork-R1V4-30B-A3B & 66.1 & -- & -- & 67.2 & 38.4 & -- & -- & -- & -- & -- & -- \\
Vision-DeepResearch-30B-A3B & 69.6 & -- & 77.6 & 74.2 & 53.7 & -- & 28.5 & \underline{37.8} & -- & -- & -- \\
REDSearcher-MM-30B-A3B\textsuperscript{$\dagger$} & 72.9 & -- & 79.3 & -- & \underline{57.2} & 23.5 & 26.6 & -- & -- & -- & -- \\
MM-DeepResearch-32B & 69.0 & 67.6 & 68.0 & 70.1 & 43.0 & -- & -- & -- & -- & -- & -- \\
LMM-Searcher-30B-A3B\textsuperscript{$\dagger,\ddagger$} & 71.0\,/\,72.3 & -- & -- & -- & -- & 22.3\,/\,\underline{30.1} & 32.9\,/\,\underline{34.8} & -- & -- & 42.0\,/\,\underline{48.3} & -- \\
OpenSearch-VL-30B-A3B & 68.7 & 74.9 & 67.4 & 73.2 & 41.1 & -- & -- & 33.5 & -- & -- & -- \\
SimpleSearch-VL-30B-A3B & \textbf{83.6} & \underline{79.6} & \underline{81.1} & \underline{79.0} & 55.9 & -- & 34.4 & -- & -- & -- & -- \\
\rowcolor{ourspink}\textbf{DeepVoyager-VL-30B-A3B} & 74.0 & \textbf{81.0} & \textbf{82.7} & \textbf{84.7} & \textbf{64.2} & \textbf{30.5} & \textbf{40.6} & \textbf{39.4} & \textbf{35.0} & \textbf{53.8} & \textbf{58.6} \\
\rowcolor{ourspink!45}\textcolor{softgreen}{\textbf{$\Delta$ vs. base}} & \textcolor{softgreen}{\textbf{+9.3}} & \textcolor{softgreen}{\textbf{+10.0}} & \textcolor{softgreen}{\textbf{+9.4}} & \textcolor{softgreen}{\textbf{+12.4}} & \textcolor{softgreen}{\textbf{+22.6}} & \textcolor{softgreen}{\textbf{+20.6}} & \textcolor{softgreen}{\textbf{+22.9}} & \textcolor{softgreen}{\textbf{+17.8}} & \textcolor{softgreen}{\textbf{+23.7}} & \textcolor{softgreen}{\textbf{+30.7}} & \textcolor{softgreen}{\textbf{+17.9}} \\
\bottomrule[1.2pt]
\end{tabular}
}
\ifdefined\mainresultinline
\end{minipage}
\else
\end{table}
\fi

\section{Experiments}

\begin{table}[t]
    \centering
    \begingroup
    \def\frameworkcomparisoninline{}
      \setlength{\tabcolsep}{4.5pt}

  \ifdefined\frameworkcomparisoninline
  \else
  \begin{table}[t]
  \fi
  \caption{Performance comparison of different multimodal search frameworks across three base models. The average is computed over four benchmarks; ``--'' denotes unreported results.}
  \label{tab:framework-comparison}
  \centering
  \footnotesize
  \resizebox{0.95\linewidth}{!}{
  \begin{tabular}{llcccc|c}
  \toprule[1.2pt]
  \textbf{Model} & \textbf{Evaluation Method} & \textbf{MMSearch} & \textbf{MM-BC} & \textbf{MMSearch+} & \textbf{VisBrowse} & \textbf{Avg.} \\
  \midrule[0.8pt]
  \textbf{GPT-5} & Direct Answer~\citep{lmmsearcher} & 33.3 & 10.3 & 19.1 & 26.0 & 22.2 \\
   & w/ Vision-DeepResearch Agent Workflow~\citep{visiondeepresearch} & 63.7 & -- & 17.2 & -- & -- \\
   & w/ LMM-Searcher Agentic Search~\citep{lmmsearcher} & 72.2 & 23.7 & 34.8 & 35.5 & 41.6 \\
   & w/ DeepVoyager-VL (Ours) & \textbf{79.7} & \textbf{49.8} & \textbf{50.8} & \textbf{59.2} & \textbf{59.8} \\
  \midrule[0.8pt]
  \textbf{Gemini-2.5-Pro} & Direct Answer~\citep{lmmsearcher} & 39.8 & 10.3 & 14.5 & 27.2 & 23.0 \\
   & w/ Vision-DeepResearch Agent Workflow~\citep{visiondeepresearch} & 69.0 & -- & 22.2 & -- & -- \\
   & w/ LMM-Searcher Agentic Search~\citep{lmmsearcher} & 66.3 & 12.1 & 28.1 & 16.0 & 30.6 \\
   & w/ DeepVoyager-VL (Ours) & \textbf{72.0} & \textbf{19.0} & \textbf{33.4} & \textbf{43.8} & \textbf{42.1} \\
  \midrule
  \textbf{Qwen3-VL-30B-A3B-Thinking} & Direct Answer~\citep{lmmsearcher} & 17.7 & 7.1 & 2.7 & 13.0 & 10.1 \\
   & w/ Vision-DeepResearch Agent Workflow~\citep{visiondeepresearch} & 53.2 & -- & 13.6 & -- & -- \\
   & w/ LMM-Searcher Agentic Search~\citep{lmmsearcher} & 62.0 & 9.8 & 14.4 & 16.0 & 25.6 \\
   & w/ DeepVoyager-VL (Ours) & \textbf{70.0} & \textbf{10.0} & \textbf{20.9} & \textbf{20.7} & \textbf{30.4} \\
  \bottomrule[1.2pt]
  \end{tabular}
  }
  \ifdefined\frameworkcomparisoninline
  \else
  \end{table}
  \fi

    \endgroup

    \vspace{1.0em}
    \begin{minipage}[t]{0.49\linewidth}
        \vspace{0pt}
        \begingroup
        \def\dataablationinline{}
        \setlength{\tabcolsep}{4.5pt}
\ifdefined\dataablationinline
\captionsetup{type=table}
\else
\begin{table}[t]
\fi
\caption{Cumulative training-data ablation: 20K multi-source trajectories followed by 7K VIL trajectories.}
\label{tab:data-ablation}
\centering
\small
\begingroup
\renewcommand{\arraystretch}{1.12}
\resizebox{\linewidth}{!}{
\begin{tabular}{l|ccc|c}
\toprule[1.2pt]
\textbf{Training Setting} & \textbf{BC-VL} & \textbf{BC-V\textsuperscript{3}} & \textbf{VisBrowse} & \textbf{Avg.} \\
\midrule
\textbf{Qwen3-VL-8B (Agentic)} & 34.6 & 7.7 & 15.4 & 19.2 \\
$+$20K multi-source trajectories & 54.4 & 26.7 & 40.8 & 40.6 \\
\textbf{$+$7K VIL trajectories} & \textbf{58.4} & \textbf{32.3} & \textbf{47.3} & \textbf{46.0} \\
\midrule
\textbf{Qwen3-VL-30B-A3B (Agentic)} & 41.6 & 11.3 & 23.1 & 25.3 \\
$+$20K multi-source trajectories & 62.2 & 28.7 & 42.6 & 44.5 \\
\textbf{$+$7K VIL trajectories} & \textbf{64.2} & \textbf{35.0} & \textbf{53.8} & \textbf{51.0} \\
\bottomrule[1.2pt]
\end{tabular}
}
\endgroup
\ifdefined\dataablationinline
\else
\end{table}
\fi

        \endgroup
    \end{minipage}
    \hfill
    \begin{minipage}[t]{0.49\linewidth}
        \vspace{0pt}
        \begingroup
        \def\frameworkablationinline{}
          \setlength{\tabcolsep}{4.5pt}
  \ifdefined\frameworkablationinline
  \captionsetup{type=table}
  \else
  \begin{table}[t]
  \fi
  \caption{DeepVoyager-VL component ablation with Qwen3.6-35B-A3B; each setting removes one component.}
  \label{tab:framework-ablation}
  \centering
  \small
  \resizebox{\linewidth}{!}{
  \begin{tabular}{lccc|c}
  \toprule[1.2pt]
  \textbf{Setting} & \textbf{FVQA} & \textbf{VDR} & \textbf{BC-V\textsuperscript{3}} & \textbf{Avg.} \\
  \midrule
  \textbf{Full DeepVoyager-VL} & \textbf{78.0} & \textbf{36.0} & \textbf{39.3} & \textbf{51.1} \\
  \midrule
  w/o Summary & 74.0 & 35.0 & 32.3 & 47.1 \\
  w/o Image Search & 77.7 & 35.4 & 32.7 & 48.6 \\
  w/o Fetch Image & 76.3 & 35.6 & 33.3 & 48.4 \\
  w/o Crop Image & 75.7 & 33.0 & 36.3 & 48.3 \\
  \bottomrule[1.2pt]
  \end{tabular}
  }
  \ifdefined\frameworkablationinline
  \else
  \end{table}
  \fi

        \endgroup
    \end{minipage}
\end{table}

\subsection{Experimental Setup}

\paragraph{Benchmarks.}
We evaluate on ten multimodal information-seeking benchmarks: MMSearch, SimpleVQA, LiveVQA, FVQA, BrowseComp-VL, MM-BrowseComp, MMSearch-Plus, VDR-Bench, BrowseComp-V\textsuperscript{3}, and VisBrowse-Bench~\citep{mmsearch,simplevqa,livevqa,mmsearchr1,webwatcher,mmbrowsecomp,mmsearchplus,vdrbench,browsecompv3,visbrowse}. They cover factual visual QA, multimodal evidence seeking, and long-horizon visual browsing; see Appendix~\ref{app:benchmarks} for details.

\paragraph{Baselines.}
We compare three groups. \emph{Direct Answer} and \emph{Agentic Workflow} use the same five off-the-shelf models---GPT-5.5, Gemini-3.1-Pro, Claude-Opus-4.6, Qwen3-VL-8B, and Qwen3-VL-30B-A3B---without tools and in our framework, respectively. \emph{Multimodal Deep Search Agents} includes MMSearch-R1, WebWatcher, DeepEyesV2, SenseNova-MARS, Skywork-R1V4, Vision-DeepResearch, REDSearcher-MM, MM-DeepResearch, POINTS-Seeker, LMM-Searcher, OpenSearch-VL, SimpleSearch-VL, and Visual-Seeker~\citep{mmsearchr1,webwatcher,deepeyesv2,sensenovamars,skyworkr1v4,visiondeepresearch,redsearcher,mmdeepresearch,pointsseeker,lmmsearcher,opensearchvl,simplesearchvl,visualseeker}. See Appendix~\ref{app:baselines} for variants and result sources.

\paragraph{Evaluation.}
For predictions generated in our experiments, we judge answer correctness with an LLM-as-judge. We adopt the judge prompt of Vision-DeepResearch~\citep{visiondeepresearch} and use Qwen3.6-35B-A3B as the judge model across all benchmarks.

\paragraph{Implementation Details.}
We SFT Qwen3-VL-8B/30B-A3B-Instruct for four epochs using 64 NVIDIA H20 GPUs, a global batch size of 64, and a peak learning rate of $2\times10^{-5}$. We update the language backbone while freezing the vision encoder and multimodal merger, and evaluate for up to 50 turns. Full configurations are provided in Appendix~\ref{app:experimental-details}.


\subsection{Main Results}

\paragraph{Overall Performance.}

As shown in Table~\ref{tab:my-table-2-test}, both DeepVoyager-VL variants substantially outperform their direct-answer and agentic-workflow baselines. For the 30B-A3B model, introducing the agentic workflow raises the average from 20.7 to 40.7 ($+20.0$), while trajectory SFT further increases it to 58.6 ($+17.9$). The 8B model follows the same progression, from 17.5 to 35.3 ($+17.8$) and then to 54.8 ($+19.5$).
Among scale-matched open-source multimodal deep-search agents, DeepVoyager-VL-30B-A3B achieves the best reported result on nine of ten benchmarks, while DeepVoyager-VL-8B leads on eight of ten. The gains are concentrated on the more search-intensive benchmarks targeted by our method. Relative to the same backbone in our agentic workflow, the 30B-A3B model improves by 10.3 points on average over the first four benchmarks, compared with 23.1 points over the six benchmarks from BrowseComp-VL through VisBrowse-Bench; the corresponding gains for the 8B model are 13.8 and 23.4 points. The largest individual improvements are $+22.6$ on BrowseComp-VL, $+23.7$ on BrowseComp-V\textsuperscript{3}, and $+30.7$ on VisBrowse-Bench. This concentration is consistent with our goal of training agents to repeatedly acquire and consume visual evidence during extended search. Despite using an open 30B-scale backbone and SFT alone, DeepVoyager-VL reaches an average of 58.6, within 6.1--8.6 points of proprietary models under the same agentic workflow.

\paragraph{Framework Comparison.}
To assess framework generality, we pair DeepVoyager-VL with three base models and compare it against reported results from Vision-DeepResearch~\citep{visiondeepresearch} and LMM-Searcher~\citep{lmmsearcher} across four benchmarks. All framework-based evaluations use a maximum of 50 interaction turns.
As shown in Table~\ref{tab:framework-comparison}, DeepVoyager-VL yields the highest reported average for all three base models. Relative to LMM-Searcher, the average gain is 18.2 for GPT-5, 11.5 for Gemini-2.5-Pro, and 4.8 for Qwen3-VL-30B-A3B-Thinking, showing that the benefit varies with the underlying base model. The largest gains appear on VisBrowse-Bench with Gemini-2.5-Pro ($+27.8$) and on MM-BrowseComp with GPT-5 ($+26.1$), where agents must locate and inspect in-page visual evidence over extended interactions. These results are consistent with the intended roles of active visual acquisition and goal-conditioned evidence compression.

\subsection{Ablation Study and Analysis}

\paragraph{Data Ablation.}
To isolate the contribution of our VIL data, we use a cumulative design. For each backbone, we compare the off-the-shelf agentic baseline, SFT on 20K multi-source trajectories from OpenSearch-VL, Vision-DeepResearch, and REDSearcher-MM, and SFT on the full 27K mixture after adding 7K synthesized vision-in-the-loop (VIL) trajectories, with all training and inference settings fixed.
As shown in Table~\ref{tab:data-ablation}, the 20K multi-source trajectories first raise the average by 21.4 points for the 8B model and 19.2 points for the 30B model. More importantly, our 7K VIL trajectories provide further gains of 5.4 and 6.5 points, respectively. Across BrowseComp-VL, BrowseComp-V\textsuperscript{3}, and VisBrowse-Bench, these gains are 4.0, 5.6, and 6.5 points for the 8B model, and 2.0, 6.3, and 11.2 points for the 30B model. This consistent pattern across scales establishes the incremental value of our VIL data beyond a strong multi-source training baseline, with the largest benefits on image-centric tasks requiring extended search.

\paragraph{Framework Ablation.}
Using Qwen3.6-35B-A3B as the base model with a maximum of 50 interaction turns, we remove Summary, Image Search, Fetch Image, and Crop Image one at a time from the full framework.
Table~\ref{tab:framework-ablation} shows that removing any component lowers the average performance by 2.5--4.0 points. Summary produces the largest average drop, while the per-benchmark results reveal complementary roles: Summary contributes most on FVQA, cropping is most important on VDR-Bench, and all four components yield clear gains on BrowseComp-V\textsuperscript{3}. DeepVoyager-VL therefore benefits jointly from visual retrieval, active observation, local visual operations, and context management.

\begin{figure}[t]
    \centering
    \begin{minipage}[t]{0.485\textwidth}
        \centering
        \includegraphics[width=\linewidth]{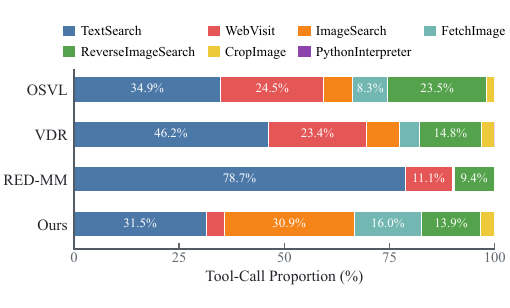}
        \\[0.25em]\textbf{(a) Visual Tool Engagement}
    \end{minipage}
    \hfill
    \begin{minipage}[t]{0.485\textwidth}
        \centering
        \includegraphics[width=\linewidth]{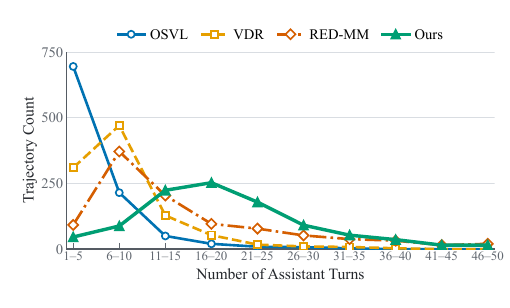}
        \\[0.25em]\textbf{(b) Interaction Horizon}
    \end{minipage}
    \caption{Trajectory statistics from unified Doubao-2.0-Pro rollouts on 1,000 examples per dataset. \textbf{(a)} Tool-call proportions by functional category. \textbf{(b)} Trajectory counts in five-turn intervals. OSVL, VDR, and RED-MM denote OpenSearch-VL, Vision-DeepResearch, and REDSearcher-MM, respectively.}
    \label{fig:trajectory-analysis}
\end{figure}

\paragraph{Trajectory Analysis.}
Beyond downstream accuracy, we compare trajectories elicited by four data sources along two complementary dimensions: vision-in-the-loop behavior and interaction horizon. We randomly sample 1,000 queries from EventVoyage-VL and each of three public multimodal search datasets: OpenSearch-VL~\citep{opensearchvl}, Vision-DeepResearch~\citep{visiondeepresearch}, and REDSearcher-MM~\citep{redsearcher}. We use Doubao-2.0-Pro to roll out every query under the same DeepVoyager-VL framework, tool environment, and inference configuration, thereby avoiding confounds from different models, tool environments, or agent implementations.

\noindent\textbf{\textit{(1) Vision-in-the-Loop Behavior.}}
We first compare the normalized tool-call composition of each data source. As shown in Figure~\ref{fig:trajectory-analysis}(a), visual tools---image search, image loading, reverse image search, and image cropping---account for \textbf{64.3\%} of all tool calls in EventVoyage-VL, compared with 40.6\% in OpenSearch-VL, 30.3\% in Vision-DeepResearch, and only 10.1\% in REDSearcher-MM. Conversely, textual search and page visiting together occupy 59.4\%, 69.7\%, and 89.9\% of the three existing datasets, respectively, but only \textbf{35.7\%} of ours. Visual evidence acquisition is therefore the dominant interaction pattern in EventVoyage-VL, consistent with the intended vision-in-the-loop behavior.

\noindent\textbf{\textit{(2) Interaction Horizon.}}
We next measure trajectory length by the number of assistant interaction turns, including the final-answer turn. Figure~\ref{fig:trajectory-analysis}(b) groups the trajectories into five-turn intervals. OpenSearch-VL peaks within 1--5 turns, while Vision-DeepResearch and REDSearcher-MM peak within 6--10 turns. In contrast, EventVoyage-VL peaks within 16--20 turns and retains a broader tail over later intervals. Its trajectories thus involve more extended sequences of evidence acquisition, visual observation, and follow-up reasoning, complementing the tool-composition analysis with evidence of a longer interaction horizon.

\section{Conclusion}

We present DeepVoyager-VL, a long-horizon multimodal deep-search method for vision-in-the-loop search. Specifically, the first step is to construct a multimodal event graph to synthesize questions with intermediate visual dependencies and long reasoning chains. We then design an agent framework for active visual acquisition and on-demand image loading, and finally fine-tune the models on the resulting trajectories without additional reinforcement learning. Across ten benchmarks, trajectory SFT improves the average performance of the 8B and 30B-A3B models by 55.2\% and 44.0\%, respectively, over the same base models in the agentic workflow. To localize these gains, data ablations show that our VIL data consistently benefit both model scales beyond a strong multi-source trajectory baseline, confirming the value of these data. Beyond training data, DeepVoyager-VL achieves the highest average performance in framework comparisons across three base models, demonstrating the framework's effectiveness and generality. Analysis of interaction behavior further shows that our data elicit a higher proportion of visual-tool calls and longer trajectories than existing open-source data, reflecting their vision-in-the-loop and long-horizon characteristics. Overall, these results establish supervised trajectory fine-tuning on data with intermediate visual dependencies as an effective path toward long-horizon multimodal search.

\clearpage
\bibliography{references}
\bibliographystyle{iclr2027_conference}


\clearpage
\appendix

\section{Related Work}
\label{app:related-work}

\subsection{Multimodal Large Language Models}

Early MLLMs connect pretrained vision encoders to language models through lightweight projection modules, establishing general visual instruction-following capabilities~\citep{llava}. Subsequent systems scale both model architectures and vision--language corpora to improve cross-modal understanding and reasoning~\citep{llavaonevision,qwen25vl,internvl3}, with proprietary systems showing similar trends at larger scale~\citep{gpt4o,gemini25}. More recent models integrate cross-modal data during pretraining and thereby inherit stronger reasoning and tool-use capabilities from their language backbones~\citep{qwen3vl,glm45v}.

Post-training research further moves beyond treating an image as a static input. In the thinking-with-images paradigm, models actively manipulate visual content to expose details needed for reasoning. One line of work uses reinforcement learning to incentivize such behavior directly~\citep{deepeyes}, while another emphasizes the importance of first establishing visual tool use through cold-start supervision. Pixel Reasoner~\citep{pixelreasoner} identifies a learning trap in which models bypass newly introduced visual tools, and Mini-o3~\citep{minio3} observes that pure reinforcement learning struggles to produce the deep visual trajectories required by difficult searches. In parallel, the available operations have expanded from fixed crops to code-synthesized image transformations~\citep{pyvision,thyme}.

These advances substantially improve how a model examines visual evidence, but they primarily operate on images already supplied with the task. The model's evidence remains bounded by its parametric knowledge and its initial visual inputs. Open-world information seeking introduces a different requirement: the model must discover previously unseen images, inspect them selectively, and use the resulting observations to determine what to retrieve next. This shift from manipulating given images to acquiring new visual evidence motivates multimodal deep-search agents.

\subsection{Multimodal Deep Search Agents}

Retrieval-augmented generation grounds model outputs in fixed external corpora~\citep{rag,dpr,murag,visrag}, whereas deep-search agents turn retrieval into an iterative process of reasoning, tool use, and evidence integration~\citep{react,searchr1,webdancer,websailor,tongyidr}. Multimodal variants extend this interaction loop with image search, reverse image search, and visual manipulation, allowing the agent to acquire evidence beyond both its parameters and the images supplied at inference~\citep{mmsearchr1,webwatcher,visiondeepresearch,opensearchvl}.

Existing systems differ in how they establish search behavior. MMSearch-R1~\citep{mmsearchr1} optimizes the policy end-to-end with an outcome-based reward and a search penalty, encouraging tool calls to be made on demand. A widely adopted alternative performs supervised cold-start training on synthesized multi-turn trajectories before reinforcement-learning refinement~\citep{webwatcher,visiondeepresearch,opensearchvl,mmdeepresearch,sensenovamars}; the pioneer experiment in DeepEyesV2~\citep{deepeyesv2} similarly shows that reinforcement learning alone does not reliably induce tool use. A third line argues that a modest collection of high-quality, planning-consistent trajectories can be sufficient, avoiding the cost and instability of an additional reinforcement-learning stage~\citep{skyworkr1v4,visualseeker}. DeepVoyager-VL follows this supervised-only route while targeting trajectories with explicit intermediate visual dependencies.

Orthogonal to the training recipe, the action space has steadily expanded. Early systems rely mainly on reverse image search, while later agents crop images to suppress background noise before retrieval~\citep{deepmmsearchr1,sensenovamars}, apply enhancement operations such as sharpening, super-resolution, and perspective correction~\citep{opensearchvl}, or execute generated code for more flexible image manipulation~\citep{skyworkr1v4}. These tools improve local perception and visual retrieval, but long-horizon interaction also requires deciding which of many candidate images deserve the cost of entering the multimodal context.

This context-management problem is coupled with evidence reliability. Search snippets, page titles, image captions, and file names are useful navigation cues but are not substitutes for answer-bearing evidence, motivating explicit verification before retrieved information is used~\citep{simplesearchvl}. Recent multimodal agents reduce context growth by offloading visual assets~\citep{lmmsearcher} or folding stale context into visual space~\citep{pointsseeker}. Related text-only approaches learn memory updates~\citep{memagent}, proactively fold context~\citep{agentfold}, or compress long histories into visual representations~\citep{glyph}. These mechanisms primarily manage evidence after it has entered the interaction history.

DeepVoyager-VL complements these approaches by managing evidence acquisition before full visual content enters the context. It separates lightweight candidate discovery from selective visual observation, and preserves acquired observations as referenceable inputs for later visual operations and retrieval. The resulting visual working memory is cumulative rather than a learned persistent memory: its purpose is to connect discovery, observation, local manipulation, and follow-up search across turns. Detailed tool interfaces and context serialization are deferred to Appendix~\ref{app:method-details}.

\subsection{Datasets and Benchmarks}

Multimodal search benchmarks increasingly place vision deeper in the interaction loop. Early datasets primarily use an input image to identify an entity before textual retrieval~\citep{mmsearch,livevqa}. Later benchmarks demand finer-grained perception of the initial visual input and stronger integration with retrieved information~\citep{mmsearchplus,vdrbench}. Recent evaluations further require agents to inspect newly discovered images, combine evidence across pages and modalities, and continue browsing from acquired visual observations~\citep{mmbrowsecomp,browsecompv3,visbrowse}. Per-benchmark task definitions and evaluation settings are provided in Appendix~\ref{app:benchmarks}; here we focus on whether existing training-data construction matches this progression.

One prevalent synthesis paradigm is \emph{entity substitution}, which first constructs a textual reasoning path and then replaces an entity mention with an image~\citep{webwatcher,opensearchvl}. Even within this paradigm, the preferred anchor position is debated. OpenSearch-VL~\citep{opensearchvl} assigns functional roles to nodes along a sampled Wikipedia path and anchors the image at the source, arguing that grounding near the answer can create single-hop shortcuts. This improves the placement of input-side visual grounding, but the image still substitutes for a node in a reasoning chain that was constructed in text.

A second paradigm performs \emph{front-loaded visual reasoning}. Vision-DeepResearch~\citep{visiondeepresearch}, for example, explores the input image and produces a detailed textual description before handing the resulting context to a text-oriented deep-research process. Visual reasoning is richer than entity substitution, but it is concentrated before the subsequent search trajectory. A third paradigm uses \emph{answer-side visual grafting}. Visual-Seeker~\citep{visualseeker} injects a subchain that retrieves an image from a fuzzy search query and extracts a visual detail, then merges this unit into an existing instance. Here visual retrieval and perception are explicit, but they are typically localized near the answer side of the original chain.

Despite their different insertion points, all three paradigms construct the principal reasoning chain in text first. Consequently, a newly acquired visual observation rarely resolves an intermediate variable that determines a later retrieval action. EventVoyage-VL instead starts from a visually enriched multimodal event graph, composes an inference program with internal visual interfaces, and only then realizes the program as a question. This structure-before-language design makes visual evidence part of the dependency structure rather than a replacement or auxiliary subchain added after textual reasoning has been formed. The construction rules and necessity checks are detailed in Appendix~\ref{app:method-details}.

\section{Method Details}
\label{app:method-details}

\subsection{Vision-in-the-Loop Data Synthesis}

\paragraph{Multimodal event graph schema.}
The synthesis pipeline represents source records, their visual evidence, and their spatio-temporal aggregation in a unified event graph. Table~\ref{tab:event-graph-schema} lists the fields exposed to subsequent program construction. Each atomic event preserves its original directed source--target relation, whereas an edge aggregates repeated interactions between an entity pair. Events sharing the same date and city are further grouped into a macro-event, which provides a local participant set from which multi-entity dependencies can be composed. Image records explicitly separate provenance and retrieval cues from visually grounded content, allowing a later program to distinguish evidence used to relocate an image from facts that must be read from its pixels.

\begin{table}[t]
\caption{Core schema of the multimodal event graph. Auxiliary bookkeeping fields are omitted.}
\label{tab:event-graph-schema}
\centering
\footnotesize
\begin{tabular}{@{}p{0.12\textwidth}p{0.44\textwidth}p{0.38\textwidth}@{}}
\toprule
\textbf{Object} & \textbf{Principal fields} & \textbf{Role in the event graph} \\
\midrule
Entity & \texttt{id}, \texttt{qid}, \texttt{type}, \texttt{label}, \texttt{aliases}, \texttt{sectors}, \texttt{countries}, \texttt{degree}, \texttt{attributes}, \texttt{images} & A person or organization together with normalized identifiers, metadata, and associated visual evidence. \\
Atomic event & \texttt{event\_id}, \texttt{date}, \texttt{src}, \texttt{tgt}, \texttt{cameo}, \texttt{cameo\_root}, \texttt{event\_text}, \texttt{intensity}, \texttt{city}, \texttt{province}, \texttt{country}, \texttt{lat}, \texttt{lon}, \texttt{story\_id}, \texttt{publisher}, \texttt{attributes}, \texttt{images} & A provenance-bearing interaction with participants, relation type, time, location, and optional visual evidence. \\
Entity-pair edge & \texttt{source}, \texttt{target}, \texttt{n\_events\_total}, \texttt{n\_dates}, \texttt{n\_years}, \texttt{years}, \texttt{first\_date}, \texttt{last\_date}, \texttt{top\_cameo\_root}, \texttt{cities}, \texttt{events} & An aggregate view of repeated interactions between two entities while retaining the constituent directed events. \\
Image & \texttt{path}, \texttt{url}, \texttt{source}, \texttt{role}, \texttt{caption}, \texttt{depicts}, \texttt{visual\_facts} & A visual asset linked to an entity or event, including its provenance, semantic role, depicted content, and pixel-grounded facts. \\
Macro-event & \texttt{cluster\_id}, \texttt{date}, \texttt{city}, \texttt{participants}, \texttt{event\_ids}, \texttt{n\_pairs} & A same-date, same-city cluster used to expose local co-occurrence structure and candidate cross-entity transitions. \\
\bottomrule
\end{tabular}
\vspace{0.8em}

\caption{Functions and input schemas of the DeepVoyager-VL tools. Conversation state required to resolve image tokens is injected by the runtime and is not a policy-visible argument.}
\label{tab:agent-tool-schema}
\begin{tabular}{@{}p{0.17\textwidth}p{0.38\textwidth}p{0.39\textwidth}@{}}
\toprule
\textbf{Tool} & \textbf{Function} & \textbf{Input schema} \\
\midrule
\textsc{TextSearch} & Retrieve up to ten title--URL--snippet candidates per query. & \texttt{query}: string or list of complementary query strings. \\
\textsc{WebVisit} & Visit pages and extract goal-relevant textual evidence and image references. & \texttt{url}: string or list of at most five URLs; \texttt{goal}: non-empty string. \\
\textsc{ImageSearch} & Discover and rank image candidates against a search goal. & \texttt{query}: string or list of at most five queries; \texttt{goal}: non-empty string. Up to five candidates are returned per query. \\
\shortstack[l]{\textsc{ReverseImage}\\\textsc{Search}} & Find visually matching images and possible source pages from an existing image. & \texttt{image\_token} and/or \texttt{image\_url}: string or list; \texttt{goal}: non-empty string. At most three images are processed per call. \\
\textsc{FetchImage} & Load selected image URLs as reusable visual observations. & \texttt{urls}: URL string, list of URL strings, or list of \texttt{\{url, caption\}} objects; at most five items. \\
\textsc{CropImage} & Inspect one or more local regions of an input or previously loaded image. & \texttt{crop\_config}: mapping from an image token to one box or a list of boxes, each \texttt{[left, top, right, bottom]} in normalized $[0,1000]$ coordinates; at most five boxes per image and ten per call. \\
\shortstack[l]{\textsc{Python}\\\textsc{Interpreter}} & Perform arithmetic, structured-data processing, and lightweight plotting in a restricted environment. & \texttt{code}: Python source string; file, network, operating-system, subprocess, and dynamic-code access are disabled. \\
\bottomrule
\end{tabular}
\end{table}

\paragraph{Construction scale.}
The source collection contains approximately 3.78 million atomic event records. Spatio-temporal aggregation produces 30,337 macro-events, from which the structure-before-language pipeline constructs approximately 7K multimodal search tasks. These counts distinguish raw event observations from the macro-event units used for program composition; the latter are not additional source events.

\subsection{Long-Horizon Multimodal Search Agent}

\paragraph{Tool interfaces.}
Table~\ref{tab:agent-tool-schema} gives the public action schema exposed to the policy. Search tools return lightweight candidates or goal-conditioned evidence, while \textsc{FetchImage} and \textsc{CropImage} are the only operations that materialize new visual observations in policy context. Batch limits bound the context and latency of a single action without preventing the agent from issuing follow-up calls.

\paragraph{Goal-conditioned summary prompt.}
Before a visited page enters the policy context, Qwen3-VL-Plus summarizes it with the current search goal. The implementation first removes navigation and boilerplate, truncates excessively long page content, and extracts up to 15 image references. The following prompt template specifies the information retained by this helper model:

\begin{quote}
\small
\textbf{System:} You are a precise multimodal-aware web content analyst. Return a valid JSON object only, with no Markdown or commentary. Never include navigation menus, sidebars, cookie banners, footers, or advertisements.

\textbf{User:} Given the \emph{webpage content}, the \emph{user goal}, and the list of \emph{images found on the page}: (1) locate the sections, tables, names, dates, numbers, and claims directly related to the goal, and explain briefly why the page is relevant; (2) extract the strongest supporting content from the article body while preserving qualifications and source claims; (3) summarize how that evidence answers or contributes to the goal without introducing unsupported facts; and (4) select only page images that may help verify or answer the goal, retaining their \texttt{alt}, \texttt{url}, and a short \texttt{caption}. If the page contains no direct evidence, return empty evidence and an empty image list. Output exactly the keys \texttt{rational}, \texttt{evidence}, \texttt{summary}, and \texttt{relevant\_images}, with no additional keys.
\end{quote}

The image-search helper uses an analogous goal-conditioned prompt to assign each candidate a relevance level and a pixel-grounded caption. Reverse image search instead compares the query and candidate pixels and reports one of four match states: visual match, partial match, not a match, or insufficient visual evidence. These helper outputs guide navigation; answer-bearing visual claims still require the agent to load and inspect the selected image explicitly.

\subsection{Data Curation and Agent Training}

\paragraph{Candidate data mixture.}
In addition to our synthesized tasks, the candidate pool includes public trajectories associated with LiveVQA~\citep{livevqa}, FVQA as released with MMSearch-R1~\citep{mmsearchr1}, WebQA~\citep{webqa}, REDSearcher-MM~\citep{redsearcher}, and Vision-DeepResearch~\citep{visiondeepresearch}. We normalize all sources to the common tool protocol in Table~\ref{tab:agent-tool-schema} and apply the filtering procedure below. The retained training mixture contains approximately 20K open-source trajectories and 7K trajectories generated from our own tasks.

\paragraph{Direct-answer filtering.}
We first test whether a candidate genuinely requires external tools. Qwen3-VL-8B-Instruct independently answers each question three times without tool access, and Qwen-3.6-Flash judges each response against the reference answer. We discard a question if any of the three attempts is correct (Pass@3), since a single successful direct answer indicates that tool use is not necessary to solve it.

\paragraph{Difficulty stratification.}
For each remaining question, Qwen3-VL-8B-Instruct produces eight tool-augmented rollouts, with correctness again determined by Qwen-3.6-Flash. Questions with 7--8 correct attempts are labeled easy and removed; those with 4--6 correct attempts are retained as medium, and those with 0--3 correct attempts are retained as hard. These rollouts are used only to estimate question difficulty: the questions are retained, but the probe trajectories themselves are never included in the supervised corpus.

\paragraph{Teacher trajectory generation and verification.}
We regenerate trajectories for all retained medium and hard questions using \texttt{doubao-seed-2-0-pro-260215} under the DeepVoyager-VL tool interface. Qwen-3.6-Flash then judges the final answer of each trajectory, and only correct trajectories are retained for training. This separation prevents the lower-capacity rollouts used for filtering and stratification from becoming supervision targets.

\paragraph{Supervised fine-tuning.}
All retained trajectories are serialized as interleaved assistant reasoning, tool calls, tool observations, and final answers. The loss mask supervises every assistant-generated reasoning token, serialized tool call, and final-answer token. System and user messages, multimodal placeholders, and all tool-return tokens remain conditioning context and do not contribute to the loss. The optimization settings and exact training configuration are reported in Appendix~\ref{app:implementation-details}.

\FloatBarrier

\section{Experimental Details}
\label{app:experimental-details}

\subsection{Experimental Setup}
\label{app:experimental-setup}
\label{app:benchmarks}
\label{app:baselines}

\paragraph{Benchmarks.}
We evaluate DeepVoyager-VL on ten public benchmarks spanning multimodal factual search, deep browsing, and visual-native information seeking. The main-result table orders them chronologically by release date; below, we group them by their primary evaluation focus.

\noindent\textit{Multimodal factual and knowledge-seeking benchmarks.}
MMSearch~\citep{mmsearch} evaluates whether a model can retrieve and integrate visual and textual web evidence. SimpleVQA~\citep{simplevqa} focuses on factual visual question answering, whereas LiveVQA~\citep{livevqa} emphasizes up-to-date visual knowledge that may not be available in model parameters. FVQA~\citep{mmsearchr1} covers diverse factual knowledge types and includes both search-required and search-free questions.

\noindent\textit{Deep multimodal browsing benchmarks.}
BrowseComp-VL~\citep{webwatcher} extends BrowseComp to vision--language queries that require multi-step browsing. MM-BrowseComp~\citep{mmbrowsecomp} evaluates complex multimodal browsing and evidence integration across multiple retrieval steps. MMSearch-Plus~\citep{mmsearchplus} extends MMSearch with stronger provenance requirements, while VDR-Bench~\citep{vdrbench} evaluates the complementary roles of visual and textual search within a unified protocol.

\noindent\textit{Visual-native and verifiable browsing benchmarks.}
BrowseComp-V\textsuperscript{3}~\citep{browsecompv3} targets visual, vertical, and verifiable browsing problems, with performance reported as Success Rate. VisBrowse-Bench~\citep{visbrowse} contains visual-native queries for which answer acquisition inherently requires inspecting and reasoning over visual web content. The remaining benchmarks are reported using their respective accuracy-based evaluation protocols.

\paragraph{Baselines.}
Following the organization of Table~\ref{tab:my-table-2-test}, we divide the comparison methods into three groups that distinguish tool-free model capability, gains from the agent framework, and performance after specialized multimodal search training.

\noindent\textit{Direct Answer.}
This setting measures the parametric knowledge and visual reasoning of general-purpose MLLMs without external tools. It includes three proprietary models---GPT-5.5, Gemini-3.1-Pro, and Claude-Opus-4.6---and two open-source models---Qwen3-VL-8B-Instruct and Qwen3-VL-30B-A3B-Instruct. Each model receives the original multimodal query and directly produces a final answer without search augmentation.

\noindent\textit{Agentic Workflow.}
This setting equips the same five general-purpose MLLMs with the DeepVoyager-VL tool environment while keeping their parameters unchanged. Relative to Direct Answer, it measures the benefit of iterative retrieval and visual tool access. Relative to the corresponding DeepVoyager-VL model in the \emph{Ours} block, it provides a same-backbone, same-framework baseline that isolates the additional contribution of trajectory supervision.

\noindent\textit{Multimodal Deep Search Agents.}
This group contains models specifically trained for multimodal retrieval and browsing: MMSearch-R1~\citep{mmsearchr1}, WebWatcher~\citep{webwatcher}, DeepEyesV2~\citep{deepeyesv2}, SenseNova-MARS~\citep{sensenovamars}, Skywork-R1V4~\citep{skyworkr1v4}, Vision-DeepResearch~\citep{visiondeepresearch}, REDSearcher-MM~\citep{redsearcher}, MM-DeepResearch~\citep{mmdeepresearch}, POINTS-Seeker~\citep{pointsseeker}, LMM-Searcher~\citep{lmmsearcher}, OpenSearch-VL~\citep{opensearchvl}, SimpleSearch-VL~\citep{simplesearchvl}, and Visual-Seeker~\citep{visualseeker}. As in the main table, we separate small- and large-scale variants to avoid conflating method design with model scale. We transcribe each method's final self-reported model---the post-RL checkpoint for methods using reinforcement learning and the final SFT or merged checkpoint otherwise. Unreported model--benchmark pairs are marked with ``--'' rather than filled from secondary sources.

\paragraph{Evaluation protocol.}
All agents evaluated in our environment are allowed at most 50 interaction turns. We assess answer correctness with the evaluation prompt adopted from Vision-DeepResearch~\citep{visiondeepresearch}, using Qwen3.6-35B-A3B as the judge across benchmarks. Results attributed to prior specialized agents are self-reported under the settings of their respective papers; we therefore preserve their reported benchmark coverage and do not impute missing entries or compute averages over incomparable subsets.

\subsection{Implementation Details}
\label{app:implementation-details}

\paragraph{Training configuration.}
We supervise Qwen3-VL-8B-Instruct and Qwen3-VL-30B-A3B-Instruct on the same final mixture of 27,180 multimodal tool-use trajectories. Both models are trained for four epochs with sequence packing and a maximum sequence length of 131,072 tokens. We update the language backbone while freezing the vision encoder and multimodal merger, and use bfloat16 precision throughout. Training is performed with Adam ($\beta_1=0.9$, $\beta_2=0.95$, $\epsilon=10^{-8}$), weight decay 0.1, and gradient clipping at 1.0. The learning rate is warmed up over the first 5\% of training to $2\times10^{-5}$ and then decayed with a cosine schedule to $5\times10^{-7}$.

Training uses a global batch size of 64 and a micro-batch size of one on 64 NVIDIA H20 GPUs across eight nodes. We use tensor parallelism of 4, context parallelism of 2, and sequence parallelism for both models; the 30B-A3B model additionally uses expert parallelism of 8. Packing produces 3,186 sequences, resulting in 199 optimizer steps over four epochs. We do not construct a separate validation split or select a checkpoint using validation performance; the final checkpoint at step 199 is used for evaluation. The model and data seeds are both set to 42.

\paragraph{Software and inference environment.}
We implement training with ms-swift 4.0.3 and Megatron-Core 0.15.3 on PyTorch 2.10.0, using FlashAttention for long-context training. At inference time, the agent uses Serper\footnote{\url{https://serper.dev}} for web and image retrieval, Jina\footnote{\url{https://jina.ai}} for webpage access, and Cloudflare\footnote{\url{https://www.cloudflare.com}} to host retrieved images that must be materialized in the multimodal context. Unless otherwise noted, all evaluations use the same tool schemas and a maximum budget of 50 turns.

\FloatBarrier

\section{Limitations and Future Work}
\label{app:limitations}

\subsection{Limitations}

\paragraph{Scope: search rather than full research.}
DeepVoyager-VL is designed and evaluated primarily for answer-oriented multimodal search. In the current setting, the agent receives a relatively well-specified question, gathers external evidence, and returns a concise answer. Accordingly, the evaluated benchmarks emphasize answer correctness and evidence acquisition rather than the complete research process. Our experiments do not systematically assess open-ended objective decomposition, iterative hypothesis formation, comparison of conflicting sources, long-form synthesis, or fine-grained citation organization. The reported results should therefore be interpreted as evidence of stronger long-horizon multimodal search, rather than as a complete evaluation of an autonomous multimodal research agent.

\paragraph{Latency from goal-conditioned summarization.}
The goal-conditioned summarization mechanism reduces the amount of raw webpage and image-search content entering the policy context, but it does not eliminate the associated computation. Calls to \textsc{WebVisit}, \textsc{ImageSearch}, and \textsc{ReverseImageSearch} invoke an auxiliary vision-language model to extract task-relevant evidence and image references. Because this processing lies on the interaction path between a tool response and the policy's next action, it introduces additional inference latency and deployment cost, particularly in trajectories involving many visited pages. Our current evaluation focuses on task accuracy and interaction behavior and does not systematically characterize the trade-off among summarization quality, context reduction, latency, and monetary cost.

\paragraph{Boundaries of the visual working memory.}
The current visual working memory preserves input images, fetched images, and cropped observations as referenceable objects throughout a trajectory. This design supports follow-up visual operations without repeatedly materializing the same asset, but the memory remains cumulative and trajectory-local. It does not learn which observations should be retained, consolidated, compressed, or discarded, nor does it explicitly organize evidence according to source, subtask, or confidence. As the interaction horizon grows, redundant or low-value observations may therefore accumulate, and relevant evidence may become increasingly difficult for the policy to locate within its context. The present framework consequently addresses evidence accessibility more directly than long-term evidence organization.

\subsection{Future Work}

\paragraph{From multimodal search to multimodal research.}
A natural extension is to move from concise-answer search toward full multimodal research. Such a setting would require the agent to translate an open-ended objective into a hierarchy of research questions, maintain and revise a plan as evidence is collected, reconcile agreement and conflict across sources, and produce a structured report with traceable citations. Evaluation would likewise need to extend beyond final-answer correctness to measure source coverage, citation completeness, evidential faithfulness, treatment of conflicting information, and the quality of long-form synthesis. This direction would test whether vision-in-the-loop interactions remain useful when visual evidence contributes not only to a final answer but also to an evolving research argument.

\paragraph{More efficient evidence processing.}
Future systems could reduce summarization overhead through adaptive invocation, caching, batching, asynchronous execution, and smaller distilled summarizers. For example, the harness could bypass auxiliary summarization for short or already structured responses, reuse cached page representations when the same source is revisited, and invoke a stronger multimodal summarizer only when image-bearing evidence is likely to affect a downstream decision. Summary generation could also be performed in parallel with other retrieval actions when no immediate dependency exists. More generally, jointly optimizing evidence utility, context consumption, and latency would provide a more realistic objective for deployed long-horizon agents than optimizing answer accuracy alone.

\paragraph{A modular research-agent harness.}
The current framework could be extended with a richer agent harness: an execution layer surrounding the policy that coordinates planning, tools, memory, and intermediate research artifacts. Such a harness could include a hierarchical multimodal memory that separates transient working observations from consolidated evidence; an evidence index for retrieving earlier text, images, and crops; a provenance graph connecting claims to sources; and learned retention and eviction policies that control context growth. Complementary modules could manage subtask scheduling, parallel tool execution, checkpointing, and recovery from failed searches. Together, these components would allow the agent to suspend and resume investigations, reorganize evidence as its plan changes, and operate over substantially longer research horizons without placing the entire interaction history in the policy context.

\end{document}